\documentclass[12pt]{arXiv}

\title[Sequential Topological Representations for Predictive Models]{Sequential Topological Representations \\ for Predictive Models of Deformable Objects}

\usepackage{url}
\usepackage{times}
\usepackage[utf8]{inputenc}
\usepackage{amsmath,amssymb,amsfonts,bbm}
\usepackage{gensymb} 
\usepackage{psfrag}
\usepackage{pstricks}
\usepackage{wrapfig}
\usepackage{float}
\usepackage{caption,setspace}
\SetCommentSty{mycommfont}
\newcommand{\algrule}[1][.2pt]{\par\vskip0.15\baselineskip\hrule height #1\par\vskip0.15\baselineskip}

\newcommand{\rps}{\operatorname{Rps}}
\newcommand{\VR}{\operatorname{VR}}

\newtheorem{asm}{Assumption}
\newtheorem{dfn}{Definition}

\newtheorem{cor}{Corollary}
\newtheorem{lmm}{Lemma}

\newcommand{\bx}{\pmb{x}}
\newcommand{\bmu}{\pmb{\mu}}
\newcommand{\bSigma}{\pmb{\Sigma}}

\newcommand{\bv}{\pmb{v}}
\newcommand{\tpk}{t\text{+}1:t\text{+}k}
\newcommand{\tmh}{t\text{-}h:t}
\newcommand{\R}{\mathbb{R}}

\definecolor{darkgreen}{RGB}{40, 150, 40}
\definecolor{gray}{RGB}{140, 140, 140}
\definecolor{darkmagenta}{HTML}{9c027b}

\makeatletter
\newcommand{\removelatexerror}{\let\@latex@error\@gobble}
\makeatother

\hypersetup{urlcolor=blue} 

\author{%
 \Name{Rika Antonova}\thanks{Rika and Anastasia contributed equally. Work was performed at KTH, supported in part by the Knut and Alice Wallenberg Foundation, and European Research Council (ERC) under the European Union's Horizon 2020 research and innovation programme (grant agreement No. 884807).} \Email{rika.antonova@stanford.edu}\\
 \addr Stanford University, Stanford, CA, USA
 \AND
 \Name{Anastasia Varava}$^{\ast}$ \Email{varava@kth.se}\\
 \Name{Peiyang Shi} \Email{pyshi@kth.se}\\
 \Name{J. Frederico Carvalho} \Email{jfpbdc@kth.se}\\
 \Name{Danica Kragic} \Email{dani@kth.se}\\
 \addr KTH Royal Institute of Technology, Stockholm, Sweden%
}

\begin{document}

\maketitle
\vspace{-30px}

\begin{abstract}
Deformable objects present a formidable challenge for robotic manipulation due to the lack of canonical low-dimensional representations and the difficulty of capturing, predicting, and controlling such objects. We construct compact topological representations to capture the state of highly deformable objects that are topologically nontrivial. We develop an approach that tracks the evolution of this topological state through time. Under several mild assumptions, we prove that the topology of the scene and its evolution can be recovered from point clouds representing the scene. Our further contribution is a method to learn predictive models that take a sequence of past point cloud observations as input and predict a sequence of topological states, conditioned on target/future control actions. Our experiments with highly deformable objects in simulation show that the proposed multistep predictive models yield more precise results than those obtained from computational topology libraries. These models can leverage patterns inferred across various objects and offer fast multistep predictions suitable for real-time applications.
\end{abstract}

\section{Introduction}

Dealing with highly deformable objects in robotics entails unique challenges. Since the shape of such objects is dynamic, canonical low-dimensional representations suitable for rigid objects mostly fail to capture the information necessary for modeling, planing and control. A black-box approach of training a large neural network (NN) to solve a particular task lacks modularity. With such approaches, new policies need to be trained for each task; moreover, these do not yield interpretable representations. Furthermore, flexible NN models that excel in capturing local features useful for control are not guaranteed to capture high-level structure needed for planning advanced tasks.

\begin{wrapfigure}{r}{0.22\textwidth}
\vspace{-5px}
\includegraphics[width=0.22\textwidth]{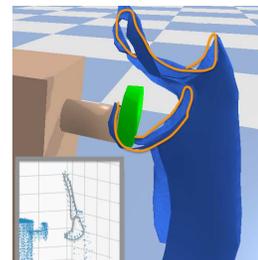}
\vspace{-20px}
\caption{Hook \& apron}
\label{fig:hook_and_apron}
\vspace{-17px}
\end{wrapfigure}

Topology can capture global shape properties of objects, such as their connectivity, holes, voids, and spacial relationships between them, while ignoring unnecessary details.
In robotic manipulation, notions and tools from topology can be used to efficiently represent scenes, objects, and their states~\citep{stork, pokorny, varava-tro, bohg}. Topological representations are especially promising for deformable objects, since many topological properties are \emph{invariant} under continuous deformations, and thus capture the shape and behaviour of objects in a succinct way.
In this work, we tackle the problem of constructing compact  topological representations for highly 
deformable objects. 
Figure~\ref{fig:hook_and_apron} illustrates one of the scenarios we consider: putting an apron on a hook. To perform such task, it is crucial to identify and control the neck strap of the apron, while representing other parts of the object explicitly might not be necessary and increases the complexity of the object model. The openings of the apron can be found as \emph{topological features} of the object point cloud without any semantic labeling; a low-dimensional topological state representation of the apron thus consists of the main openings, their location and width. 

We propose a rigorous formulation for extracting topological state representations and analyzing their evolution over time. Under several assumptions about system dynamics and observation quality, we prove that it is possible to detect significant topological features (such as the straps of an apron) and observe their dynamics from point clouds without any semantic labels. We propose the \emph{sequential persistent homology} algorithm (seqPH), building upon the persistent homology framework~\citep{edelsbrunner-tda} that can infer the topology of static point clouds.
We validate the proposed algorithm on simulated scenarios with clothing items and flexible bags.

The proposed seqPH algorithm is directly applicable to settings that use the extracted representations in an offline manner. To make our method suitable for real-time planning and control we propose to learn predictive NN models. These models take a sequence of point cloud observations as input, and output predictions for the relevant topological features up to a horizon of $k$ steps, conditioned on a future/desired sequence of control actions.
The resulting multistep predictive models
are well-suited for real-time planning and control. Since the proposed topological features are interpretable, various task-specific objective/cost functions can be obtained by using topological tools, such as the linking number -- a topological invariant that captures the `linking' or `entanglement' relationship between two curves~\citep{ho2009character, pokorny, varava-tro}. For instance, to hang an apron on a hook we would want to achieve a spatial relationship between the hook and the apron openings, which can be described with the linking number. These features are general
and can be used for more complex manipulation scenarios, such as knot tying~\citep{bohg} and assistive dressing~\citep{tamei2011reinforcement}.

Overall, our contributions yield a theoretically rigorous approach for tracking topological state and a scalable learning component that enables future applications to real-time planning \& control.

\vspace{-10px}
\section{Related Work in Robotic Manipulation and Learning Predictive Models}

Topological representations have been used in robotic manipulation for various purposes. \cite{stork, pokorny} propose a method for grasping rigid objects with `holes' (non-trivial first homology group), such as door handles and cups. In~\cite{varava-tro}, the concept of linking number is used to compute caging grasps for objects with narrow parts. In~\cite{bohg}, `topological motion primitives' (based on the linking number) are used for tying knots.
In this work, we consider one of the most challenging classes of objects: highly deformable objects with nontrivial topology, such as clothing items and flexible bags. The problem of sensing and manipulation with such objects is a part of the more general topic on deformable object manipulation. Recent surveys and benchmarks summarize the relevant scenarios and approaches:~\cite{sanchez2018robotic, garcia2020benchmarking}. Significant progress has been achieved for tasks such as flattening, spreading and folding, e.g.~\cite{van2010gravity, miller2012geometric, lakshmanan2013constraint, doumanoglou2016, nair2017combining, mcconachie2020interleaving, li2018model, lippi2020latent}. However, many of these works either consider topologically trivial objects (tablecloth, cloth patches, ropes) or lay the objects flat of the table and do not exploit the more complex aspects of their shapes (e.g. loops and holes, cylindrical parts). 
More relevant to our work are the sub-tasks considered in assistive dressing. \cite{ho2009character} addressed the motion synthesis problem by constructing a succinct topological representation. \cite{tamei2011reinforcement} used it to create a reward/cost function for optimizing a policy of putting a T-shirt on a mannequin. However, this required an accurate motion capture system to track markers for the T-shirt neck and sleeve loops. Recent works aimed to leverage
simulators, as in \cite{clegg2018learning} with the task of animating character dressing, and in \cite{yu2017haptic} with putting on a sleeve in simulation and estimating parameters to close the sim-to-real gap. 
However, these works did not resolve the problem of constructing low-dimensional representations for deformable objects. 

\vspace{5px}

One-step forward models $p(X_{t\text{+}1}|X_t,U_t)$ have been used ubiquitously in control and model-based reinforcement learning~\citep{deisenroth2013survey, moerland2020model}. However, these assume the state to be fully observable.
Predictive State Representations (PSRs)~\citep{littman2001predictive} aimed to address partial observability, but either required simplifying assumptions or were computationally expensive. Nonetheless, PSRs have been used successfully in several areas of robotics~\citep{boots2013hilbert, stork2015learning}.
\cite{hefny2018recurrent} proposed to encode PSR-like states into recurrent NNs, but training RNNs is non-trivial and this work has yet to be applied to large-scale settings. Recent works proposed adding an objective to reconstruct future states when learning to encode history of observations into lower-dimensional latent states~\citep{yin2017hashing, zintgraf2019varibad}. 
However, these lack theoretical guarantees regarding what is captured in the latent states and do not support incorporating any structured domain knowledge or representations.

In this work, instead of learning lower-dimensional representations in an unsupervised black-box way, we construct predictive models for interpretable low-dimensional states. We take the middle way between partially observable approaches like PSRs and fully observable single-step forward models. Our models have a fixed size of history and prediction horizons. This allows us to employ NN architectures that offer fast and stable training.

\vspace{-2px}
\section{Topology Background: Definitions and Notation}
\label{topo-background}

\begin{figure}
    \centering
    \includegraphics[width=0.98\textwidth]{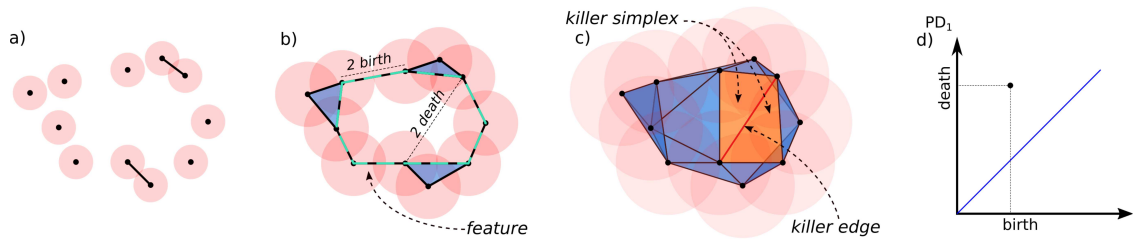}
    \caption{Construction of a Vietoris-Rips complex with growing filtration values. a) Starting from a set of points in black at each filtration level $r$ an edge is drawn between any pair of points that is at a distance $<\!2r$. Any clique of $k$ vertices in the resulting graph gives rise to a $k\!-\!1$ simplex in the resulting complex. b) When a topological feature (a loop) is formed by adding an edge of length $2x$, we say that the feature has birth-time equal to $x$. c) The loop dies when it is filled in by 2-dimensional simplices, which occurs when the red edge and consequently its two adjacent orange triangles are added. d) Persistence diagram: death versus birth. }
    \label{fig:vietoris-rips}
\vspace{-10px}
\end{figure}

Here, we briefly describe the necessary definitions from computational topology (see~\cite{koplik} for an informal introduction with animations and~\cite{edelsbrunner-tda} for a formal introduction).
Given a point cloud observation, we want to recover the topology of the underlying scene. A \emph{simplicial complex} consists of simplices and provides a way to discretely represent a topological space (the scene, in our case). A $k$-dimensional \emph{simplex} $\sigma$ can be defined as a convex hull of $k+1$ points: a single point (or a vertex) is a 0-dimensional simplex, a segment (or an edge) is a 1-dimensional simplex, a triangle (or a face) is a 2-dimensional simplex, etc. The diameter of a simplex $\sigma$ is the maximum distance between any 2 points in $\sigma$, and is $0$ in case $\sigma$ is a single point.  We use a special kind of simplicial complexes:

\allowdisplaybreaks
\begin{dfn}[Vietoris-Rips complex]
Consider a finite set of points $P \subset \mathbb{R}^n$  and  $r > 0$.  The \emph{Vietoris-Rips filtration} $\rps(\sigma)$ is a function of a simplex that is equal to half of its diameter:  $\rps(\sigma) = 1/2 \max_{p_i, p_j \in \sigma} d(p_i, p_j)$. The Vietoris-Rips simplicial complex $\VR_r(P)$  consists of all simplicies $\sigma$ such that $\rps(\sigma)$ is less than or equal
to $r$.
\end{dfn}

\noindent 
Given $r > 0$, a Vietoris-Rips complex $\VR_r(P)$ can represent a union of balls $B_r(P)$ of radius $r$ centered at the points $P$: a 0-dimensional simplex is any point from $P$, a 1-dimensional simplex is any segment between 2 points from $P$ such that the respective balls overlap, a 2-dimensional simplex is any triangle formed by 3 points from $P$ such that all 3 balls overlap pair-wisely, etc. $\VR_r(P)$ exhibits $k$-dimensional \emph{topological features}, formally referred to as $k$-dimensional homology classes: connected components ($k=0$), holes/loops ($k=1$), voids ($k=2$), etc. A Vietoris-Rips filtration can be seen as growing the radius $r$ of the balls and considering the respective complexes $\VR_r(P)$. For any $r' > r$, we will have $\VR_r(P) \subseteq \VR_{r'}(P)$. As $r$ grows, the topology of $\VR_r(P)$ changes: connected components merge together, holes appear and then get filled (formally, features become trivial), see Figure~\ref{fig:vietoris-rips}. \emph{Persistent homology} can be used to find topological features that remain for different values of $r$, and thus describe the underlying topology of the space approximated by $P$.

\begin{dfn}[Topological feature]  A \emph{topological feature} of a point set $P$ is a non-trivial homology class of $\VR_r(P)$ for some $r$.
A $k$-dimensional feature $f$ has a \emph{birth value} $birth(f)$ equal to the smallest filtration value $r$ at which it appears in $\VR_r(P)$.
Similarly, $f$ has a \emph{death value} $death(f)$ equal to the filtration value at which $f$ is trivial in $\VR_r(P)$.
The \emph{lifetime} of a feature $f$ is the difference between these values, $LT(f) = death(f) - birth(f)$.
The lifetime information can be collected in a \emph{persistence diagram}, denoted $PD_{k}(P)$ which is a set of the form $\{(f, birth(f), death(f))\}$.
\end{dfn}

\noindent 
The lifetime of a feature indicates its significance: features that die soon after being born are likely to be present due to noise in the point cloud, while features with high lifetime values are likely to be present in the underlying true space. In our case, loops with high lifetime correspond to handles and openings of objects. A set of simplices comprising a feature $f$ is called a \emph{representative} $\hat{f}$ of $f$. Each feature can have multiple representatives. Two representatives of the same feature are called \emph{homologous}.
The death of a feature occurs when a certain filtration value is passed and the interior of the feature gets filled in by a simplex, or, formally, becomes trivial (for example, in dimension~1: a hole is filled in by triangles). This leads us to consider the simplices that ``kill'' the feature.
In the Vietoris-Rips case, a simplex is added to the filtration as soon as all its edges are added. Thus, the longest edge is added together with the higher-dimensional simplices adjacent to it. Hence, this edge ``kills'' the feature.

\begin{dfn}[Killer edge]
Let $f$ be a $k$-dimensional topological feature, then the \emph{killer edge} $\sigma$ of $f$ is the edge with $\rps_{death(f)}(P)$ that leads to the death of $f$. Similarly, the \emph{killer simplices} of $f$ are the simplices of dimension $k+1$ that are added to $\VR_{death(f)}(P)$ and lead to the death of $f$. 
\end{dfn}

\section{Sequential Topological State Representations}
\label{sec:seq_ph}

\vspace{-3px}
We now present our theoretical formulation for identifying topological features of the scene and their evolution over time. Let $S_t \subset \mathbb{R}^3$ be the state of the scene at time $t$. We cannot observe $S_t$ directly, and instead rely on a point cloud observation -- a finite set of points $O_t \subseteq S_t$. Under several assumptions about the quality of the  observations and the changes between states, we prove that it is possible to recover  the topological features of $O_t \subseteq S_t$. Based on this, we design an algorithm for identifying 1-dimensional topological features (loops) in a sequence of observations $O_1, ..., O_T$.
To guarantee that the significant topological features of the scene can be recovered, we assume that the observed point cloud covers $S_t$ densely enough: for each point in $S_t$ there is at least one point in $O_t$ that is $\alpha-$close to it (Assumption~\ref{asm:samples-cover}). Given this, Lemma~\ref{lmm:alpha-approximation} and Lemma~\ref{lmm:persistence-static} show that significant features (those with lifetime higher than $\alpha)$ of $S_t$ can be recovered from the observation $O_t$. 
\begin{asm}[Observation quality]
\label{asm:samples-cover}
    The space $S_t$ is covered by an $\alpha$-neighborhood of the sampling $O_t$, i.e. $S_t \subseteq B_\alpha(O_t)$, where $B_\alpha(O_t) = \bigcup_{x \in O_t} B_\alpha(x)$.
\end{asm}

\vspace{-10px}
\begin{lmm}[Feature approximation]
\label{lmm:alpha-approximation}
For every $k$-dimensional feature $f$ in $\VR_{r}(S_t)$ with $k > 0$, there exists some feature $f^O$ in $\VR_{r + \alpha}(O_t)$ so that $f^O \sim f$ (are homologous) in $\VR_{r+\alpha}(S_t)$.
\end{lmm}
\vspace{-3px}
\begin{proof}
Note that since $S_t \subseteq B_\alpha(O_t)$, $\VR_{r}(S_t)\subseteq \VR_r(B_\alpha(O_t))$, which means that for every simplex $\sigma $ in $\hat f$, a representative of $f$ in $\VR_r(O_t)$, $\sigma \in \VR_r(B_\alpha(O_t))$. Now let us consider the case when $k = 1$.
In this case we can construct $f^O$ in a straight-forward manner, namely for each edge $\{p, q\}$, let $o, o'$ be the nearest neighbours to $p$ and $q$, respectively. The we can place the edge $\{o, o'\}$ in $f^O$ (if $o=o'$ there is no edge added).
Note that by construction, since $\{p, q\}\in\VR_{r}(S_t)$, then $\|p - q\| \leq 2r$, and since $p\in B_\alpha(o)$, and $q\in B_\alpha(o')$ $\|o - o'\| \leq \|o - p\| + \|p - q\| + \|q - o'\| \leq 2(r + \alpha)$, and therefore $\{o, o'\}\in \VR_{r + \alpha}(O_t)$.
The same construction follows for $k > 1$ from the fact that a simplex of dimension $k$ is in $\VR_{r+\alpha}(O_t)$ if all its edges are.   A proof that $f^O\sim f$ in $\VR_{r+\alpha}(S_t)$ can be done directly using the construction, for which we provide a sketch.
    For each simplex $\sigma = \{p_0, \ldots, p_k\}$ of a feature $f$, consider the transformed simplex $\sigma^O = \{q_0, \ldots, q_k\}$.
    Consider the simplices $\hat\sigma_i = \{ p_0,\ldots, p_i, q_i, \ldots, q_k\}$ for $i = 0, \ldots, k$.
    The union $\bigcup_{\sigma\in f}\bigcup_{i = 0}^k\hat\sigma_i$ forms a subset $\VR_{r+\alpha}(S_t)$ whose boundary is the union of $f$ and $f^O$.
\end{proof}

\vspace{-10px}
\begin{lmm}[Observability of significant features]
\label{lmm:persistence-static}
Let $O_t$ be a point cloud approximation of $S_t$. Any topological feature $f$ in $S_t$ whose lifetime is higher than $\alpha
$ has a corresponding topological feature $f^O$ in $PD_{k}(O_t)$, s.t. 
$
0 \leq birth_O(f^O) - birth_S(f) \leq\! \alpha,$
$
0 \leq death_O(f^O) - death_S(f) \leq\! \alpha$.
\end{lmm}

\vspace{-5px}
\begin{proof}
First note that $O_t\subseteq S_t$ therefore $birth_O(f^O) \geq birth_S(f)$, and $death_O(f^O) \geq death_S(f)$ which implies the left-hand sides of both inequalities.
By Lemma~\ref{lmm:alpha-approximation} given any representative $\hat f$ of $f$ in $\VR_{birth_S(f)}(S_t)$, it can be approximated by $\hat f^O$ in $\VR_{birth_S(f)+\alpha}(O_t)$ so that $\hat f \sim \hat f^O$ in $\VR_{birth_S(f)+\alpha}(O_t)$.
Since $LT_S(f) > \alpha$, $\hat f^O$ is non-trivial in $\VR_{birth_S(f) + \alpha}(S_t)$ and consequently in $\VR_{birth_S(f)+\alpha}(O_t)$. This implies that $birth_O(f^O) \leq birth_S(f) + \alpha$ which completes the first inequality.
Recall that $\VR_{death_S(f)}(S_t) \subseteq \VR_{death_S(f)}(B_\alpha(O_t)) \stackrel{\phi}{\cong} \VR_{death_S(f)+\alpha}(O_t)$,
where the map $\phi$ is given by $\sigma\mapsto \sigma^O$.
 Since $death_S(f) > birth_O(f^O)$, $\hat f^O$ is in $\VR_{death_S(f)}(S_t)$, where it satisfies $\hat f^O \sim \hat f$ which is trivial in $\VR_{death_S(f)}(S_t)$. Thus $\hat f^O$ is also trivial in $\VR_{death_S(f)}(B_\alpha(O_t))$.
Since $\phi$ preserves trivial classes, and acts as the identity on $\hat f^O$, it must be that $\hat f^O$ is trivial in $\VR_{death_S(f)+\alpha}(O_t)$.
Hence: $death_O(f^O) \leq death_S(f) + \alpha$, completing the second inequality.
\end{proof}

\noindent
Next, we analyze how the topology of the states $S_t$ changes over time. We assume that between successive time steps the state $S_t$ undergoes a transformation $\tau$ which is small enough, so the displacement of each point is bounded (Assumption~\ref{asm:tau-regularity}). With this, we can guarantee that significant topological features are preserved between consecutive states $S_t$ and $S_{t+1}$, and their lifetime does not change drastically, meaning that wide loops do not suddenly appear or collapse (Lemma~\ref{lmm:persistence-dynamic}).

\begin{asm}[Regularity of motion]
\label{asm:tau-regularity}
\thickmuskip=0.5\thickmuskip
    \hspace{-5px}Every $x\in S_{t}$ satisfies $\|\tau(x) - x\| < \varepsilon$, for some $\varepsilon < \tfrac{1}{2} \alpha$.
\end{asm}
 
\vspace{-10px}

\begin{lmm}[Temporal persistence of topological features]
\label{lmm:persistence-dynamic}
Consider two consecutive states $S_t$ and $S_{t+1}$. Any topological feature $f$ in $PD_{k}(S_{t})$ with lifetime higher than $ \varepsilon$ has a corresponding feature in $PD_{k}(S_{t+1})$, such that: 
$|birth_{S_t}(f) - birth_{S_{t+1}}(f) |\leq  \varepsilon,$ 
$\ |death_{S_t}(f) - death_{S_{t+1}}(f) |\leq  \varepsilon$.
\end{lmm}
\vspace{-5px}
\begin{proof}
Since $S_t$ and $S_{t+1}$ have the same set of points, $\tau$ only affects the distance between them.
From Assumption~\ref{asm:tau-regularity}, we know that $\tau(\VR_r(S_t)) \subset \VR_{r+\varepsilon}(S_{t+1})$ for all $r \geq 0$ therefore, given any representative $\hat f$ of a topological feature in $\VR_r(S_t)$, it is necessarily the case that $\tau(\hat f)$ is a representative of the same topological feature in $\VR_{r+\varepsilon}(S_{t+1})$.
Furthermore, since the birth and death of this representative corresponds to half the length of some edge which has the property of being the longest edge in some finite set of edges, this length can change by at most $2\varepsilon$ and therefore we have:
$|birth_{S_t}(\hat f) - birth_{S_{t+1}}(\tau (\hat f))| \leq  \varepsilon,$
$\ |death_{S_t}(\hat f) - death_{S_{t+1}}(\tau (\hat f))| \leq  \varepsilon$. 
Since this is true for all representatives, it is true for the feature $f$ (with $\hat f$ as a representative of $f$).
\end{proof}

\noindent We have shown that, under Assumptions~\ref{asm:samples-cover} and~\ref{asm:tau-regularity}, the topology of the scene can be recovered from observations, and, moreover, does not drastically change over time. In theory, this makes it possible to track the topological features. In practice, however, we can have several topological features with similar lifetime, and to distinguish them we need additional information about their \emph{geometric} location.
For this, we will identify each feature with the corresponding killer edge, and observe how it moves over time. Thus, we assume that a killer edge representing each feature can be uniquely identified. Furthermore, we assume that different topological features in $S_t$ are located far enough from each other, so it is possible to distinguish their respective killer edges based on Hausdorff distance $d_H(.,.)$ (Assumption~\ref{asm:uniqueness}). Then, Lemma~\ref{lmm:observation_matching} shows that killer edges from $S_t$ can be recovered given an observation $O_t$. Lemma~\ref{lmm:regularity_features} shows that the motion of killer edges is limited between consecutive states $S_t$ and $S_{t+1}$, and thus we can track them over time given the respective observations $O_t$ and $O_{t+1}$: if two killer edges in consecutive observations are close enough to each other, then they correspond to the same topological feature in $S_t$ and $S_{t+1}$ (Corollary~\ref{cor:motion_matching}).

\begin{asm}[Uniqueness of killer edges and feature separation]
\label{asm:uniqueness}
    For each $S_t$ and each feature $f$ in $PD_k(S_t)$ with $LT(f) > \alpha$, there is a unique edge $\sigma_{kill}(f)$ that kills $f$, and any other edge $\sigma'$ with filtration value satisfying $|\rps(\sigma_{kill}(f)) - \rps(\sigma')| \leq \alpha$ is $\beta$-close to $\sigma_{kill}(f)$: $d_H(\sigma', \sigma_{kill}(f)) \leq \beta$. Furthermore, any 2 distinct features $f_1, f_2$ in $PD_{k}(S_t)$ are separated: \\
$d_H(\sigma_{kill}(f_1),\sigma_{kill}(f_2)) > 2\alpha + \beta + \varepsilon$.
\end{asm} %

\vspace{-5px}

\begin{lmm}[Recovering killer  edges from observations]
\label{lmm:observation_matching}
Consider a  feature $f^S \in PD_{k}(S_t)$, for any $k \geq 1$. There exists a corresponding feature $f^O \in PD_{k}(O_t)$ s.t. $d_H(\sigma_{kill}(f^S), \sigma_{kill}(f^O)) \leq \beta$.
\end{lmm}
\vspace{-5px}
\begin{proof}
First, $f^O \in PD_{k}(O_t)$ exists by Lemma~\ref{lmm:persistence-static}, and $0 \leq death_O(f^O) - death_S(f^S) \leq \alpha$. Let $\sigma_{kill}(f^O) \in VR_{death_O(f^O)}(O_t)$ be the killer edge of $f^O$ in $O_t$. Then, $\rps(\sigma_{kill}(f^O)) \!=\! death_O(f^O) \!\leq\! death_S(f^S) + \alpha$. By Assumption~\ref{asm:uniqueness}, $d_H(\sigma_{kill}(f^S), \sigma_{kill}(f^O)) \!\leq\! \beta$.
\end{proof}

\vspace{-10px}
\begin{lmm}[Regularity of motion for killer edges]
\label{lmm:regularity_features}
    Let $\sigma_{kill}(f)$ be the killer edge of a feature $f \in PD_k(S_t)$. Its Hausdorff distance to $\sigma_{kill}(\tau (f))$ that kills its image $\tau(f)$ does not exceed $\beta + \varepsilon$.
\end{lmm}
\vspace{-5px}
\allowdisplaybreaks
\begin{proof}
By Lemma~\ref{lmm:persistence-dynamic}, we have  $|\rps(\sigma_{kill}(f)) - \rps(\sigma_{kill}(\tau(f)))| \leq \varepsilon$, where $\sigma_{kill}(\tau(f))$ is the killer edge corresponding to $\tau(f)$, which is the feature in $PD_k(S_{t+1})$ corresponding to $f$.
Now, consider $\tau(\sigma_{kill}(f))$ -- the image of $\sigma_{kill}(f)$ in $S_{t+1}$. 
By Assumption~\ref{asm:tau-regularity}, the distance from each vertex of  $\sigma_{kill}(f)$ to its image $\tau(\sigma_{kill}(f))$ does not exceed $\varepsilon$, implying $|\rps(\sigma_{kill}(f)) - \rps(\tau(\sigma_{kill}(f)))| \leq \varepsilon$, and hence $|\rps(\sigma_{kill}(\tau(f))) - \rps(\tau(\sigma_{kill}(f)))| \leq 2 \varepsilon$. Furthermore,  $d_H(\sigma_{kill}(f), \tau(\sigma_{kill}(f))) \leq \varepsilon$. Finally, in Assumption~\ref{asm:tau-regularity} we established that $2 \varepsilon < \alpha$ whereby Assumption~\ref{asm:uniqueness} allows us to conclude that: \\
$d_H(\sigma_{kill}(f), \sigma_{kill}(\tau(f))) \leq d_H(\sigma_{kill}(f), \tau(\sigma_{kill}(f))) + d_H(\tau(\sigma_{kill}(f)), \sigma_{kill}(\tau(f))) \leq \varepsilon + \beta $.
\end{proof}
\vspace{-12px}

\begin{cor}[Tracking killer edges]
\label{cor:motion_matching}
Consider two consecutive observations, $O_t$ and $O_{t+1}$, and two features $f_o \in PD_k(O_t)$ and  $f'_o \in PD_k(O_{t+1})$. 
Let $f_s$ and $f'_s$ be the features in $PD_k(S_t)$ and $PD_k(S_{t+1})$, corresponding to $f_o$ and $f'_o$ respectively. If $f_s$ and $f_s'$ represent the same feature, then $d_H(\sigma_{kill}(f_o),\sigma_{kill}(f_o'))\le 2\alpha+ \beta+ \varepsilon$.
\end{cor}
\vspace{-5px}
\begin{proof}
By triangle inequality we have: 
$d_H(\sigma_{kill}(f_o),\sigma_{kill}(f_o')) \leq d_H^{f_o,f_s,f_s',f_o'}$, with
\\
$d_H^{f_o,f_s,f_s',f_o'} =
d_H(\sigma_{kill}(f_o),\sigma_{kill}(f_s))
+
d_H(\sigma_{kill}(f_s),\sigma_{kill}(f_s'))
+
d_H(\sigma_{kill}(f_s'),\sigma_{kill}(f_o'))$.
\\
By Lemma~\ref{lmm:observation_matching} $d_H(\sigma_{kill}(f_s'),\sigma_{kill}(f_o'))$ and $d_H(\sigma_{kill}(f_s),\sigma_{kill}(f_o))$ are both smaller than $\alpha$, and by Lemma~\ref{lmm:regularity_features} $d_H(\sigma_{kill}(f_s),\sigma_{kill}(f_s')) \le \beta + \varepsilon$. And so $d_H(\sigma_{kill}(f_o),\sigma_{kill}(f_o')) \leq 2\alpha + \beta + \varepsilon$.
\end{proof}

\noindent
\textbf{Algorithmic Procedure} (Algorithm~\ref{alg:topo_tracking}): Given an observation $O_t$, we compute a Vietoris-Rips filtration and 1D topological features (loops).  $PH_t$ represents a 1D persistent diagram of $O_t$, and contains 1D topological features together with their birth and death values. We filter out those whose lifetime is smaller than $\alpha$, as they are likely to appear due to noise. For each remaining loop, we extract the corresponding killer edge and its immediate neighborhood -- the two killer triangles and the triangles adjacent to them. Since killer triangles capture the geometric properties of the loop better than a killer edge,  in practice, we use them to identify and visualize the loops. The filtration value of a killer triangle corresponds to the radius of the widest part of the loop, see Figure~\ref{fig:vietoris-rips}. The list $\{\zeta_l\}_{l=1}^{L_t}$ stores a representation $\zeta_l$ for each  loop loop $l$ in $PH_t$. $\zeta_l$ contains the killer triangles (for loop $l$), their filtration values and immediate neighborhoods (adjacent triangles); $ID_l$ identifies each loop and is used for tracking (IDs are set arbitrarily in the first iteration). We use $X_t := \{\zeta_l\}_{l=1}^{L_t}$ to denote \emph{a topological state}. The matrix $Dist$ stores Hausdorff distances between the killer edges of $O_{t-1}$ and $O_t$. Since only those pairs of loops whose birth and death values are similar can correspond to the same topological feature (Corollary~\ref{cor:motion_matching}), we set the distance between them to infinity for other pairs. Given the distance matrix $Dist$, we find the best matching between the loops from $O_t$ and $O_{t-1}$ using the Hungarian algorithm~\citep{kuhn1955hungarian}. This matching, together with the IDs of the loops from the previous time step, provides a consistent labeling of features through time.

\begin{algorithm}
\small
\SetAlgoNoEnd
\DontPrintSemicolon
\SetAlgorithmName{\hspace{-8px}Algorithm}{algorithm}{List of Algorithms}
\algrule
\caption{seqPH}
\label{alg:topo_tracking}
\algrule
{\bf Input:} Sequence of observations $\mathcal{O} = \{O_1, ..., O_T\}$, parameters $\alpha, \beta$ and $\varepsilon$\\
{\bf Output:} Sequence $\{X_1, ..., X_T \}$ of topological states\\ 
\For{$O_t \in \mathcal{O}$}
{
    $PH_t =$ persistent-homology-1D($O_t$) \tcp{compute loops}
    $PH_t.$remove-small-features($\alpha$) \tcp{remove loops with lifetime $< \alpha$} 
    $L_t = |PH_t|$ \tcp{number of loops in the scene}
    $\{\zeta_l\}_{l=1}^{L_t} = $  killer-triangles($PH_t$) \tcp{extract loop representation $\zeta_l$ for each loop $l$ in $PH_t$}
    $X_t =  \{\zeta_l\}_{l=1}^{L_t}$ \tcp{new topological state}

    $Dist_{i,j} = \infty$            \tcp{initialize distances} 
    \For{ $l \in \{0, ..., L_t\}$}
    {
        \For{$l' \in \{0, ..., L_{t-1}\}$}
        {
            \If{$|\zeta_{l'}.birth - \zeta_{l}.birth| < 2\alpha+\varepsilon$ {\bf and } $|\zeta_{l'}.death - \zeta_{l}.death|  < 2\alpha+\varepsilon$}
            {
                $dist$ = Hausdorff\big($\zeta_{l}$.killer, $\zeta_{l'}$.killer\big)
                
                \lIf{  $ dist < 2\alpha + \beta + \varepsilon$ }
                {$Dist_{l, l'}$ = 
                 $Dist_{l, l'} \!=\! dist$
                 {\color{gray}// Corollary~\ref{cor:motion_matching}}
                 }
            }
        }
        $\{ID_l\}_{l=1}^{L_t} $ = matching($X_{t}$, $X_{t-1}$, $Dist$)    \tcp{matching killer triangles}
       $X_{t}$.update-loop-IDs($\{ID_l\}_{l=1}^{L_t} $) \tcp{matching new loop IDs with loops from previous scene}
    }
}
\Return{$\{X_1, ..., X_T \}$}
\algrule
\vspace{-15px}
\end{algorithm}
\section{Predictive Models for Deformable Objects}
\label{sec:pred_models}

Using the proposed sequential persistent homology algorithm we gain ability to extract compact topological representations of deformable objects. Such representations can be directly useful for any setting that requires a dataset (modeling, supervised learning, offline planning, etc). However, the best-performing computational geometry libraries still take from 100 milliseconds to several seconds per point cloud. Hence, techniques based on computational topology alone would not be feasible for real-time planning and control.
One solution could be to learn a forward model $p(O_{t\text{+}1}|O_t,U_t)$, which predicts the next high-dimensional state $O_{t\text{+}1}$ given  the current observation $O_t$ and the vector of control actions $U_t$. However, predicting point clouds directly with high precision would be an extremely challenging problem. An alternative is to employ an approach similar to filtering or any other approach that infers latent space dynamics $p(X_{t\text{+}1} | O_t, U_t)$. While such approaches can successfully model rigid object dynamics, it would be challenging to train these to be highly precise for deformable objects. Hence, planning could be ineffective for longer horizons due to error accumulation when such a single-step model is used sequentially to obtain multistep predictions.
Therefore, we propose to learn a multistep predictive model. Requiring the model to predict into the future has been shown to enhance ability to capture non-trivial patterns~\citep{guo2020bootstrap}. Moreover, by obtaining $k$ predictions from a single forward pass we can avoid a costlier alternative of advancing the model step-by-step for estimating rewards/costs of a multistep trajectory.

\vspace{-5px}
\subsection{Learning Multistep Predictive Models}

For dealing with deformable objects, we propose to learn multi-step predictive models that take point clouds of the scene as input and predict the evolution of the topological state up to a horizon of $k$ steps into the future. We construct a dataset of trajectories of point clouds and the corresponding topological states, which are computed using the algorithm from Section~\ref{sec:seq_ph} to extract 1-dimensional homologies (i.e. loops). We refer to this approach as seqPH. We then train a predictive model that takes $h$ previous point cloud observations $O_{\tmh}$ and future/target control actions $U_{\tpk}$ as input and outputs a sequence of predicted topological states for the next $k$ time steps: $X_{\tpk}$.

Each $X_t \!:=\! \{\zeta_l\}_{l=1}^{L_t}$ contains a list of topological features $\zeta_l$ for each loop $l$ identified by seqPH. $\zeta_l$ is comprised of a killer simplex/triangle and its neighboring simplices/triangles, the ID \& lifetime of the loop and the Hausdorff distance to the corresponding loop at the previous timestep (see Section~\ref{sec:seq_ph}). We train a neural network $f_{\text{NN}}(O_{\tmh},U_{\tpk})$ to produce $X_{\tpk}$ as output. The supervised training regresses directly on the topological features using an $L_2$ loss. To create fixed-size NN inputs \& outputs we pad (or sub-sample) the point clouds and fix the maximum number of loops $L_t$.
For experiments with fully-connected NNs we used 4 hidden layers (512, 512, 256, 128 units). To leverage recently proposed scalable architectures for point cloud processing, we also experimented with an alternative of first passing each input $O_t$ trough a PointNet~\citep{qi2017pointnet}.

\vspace{-3px}
\subsection{Probabilistic Interpretation of Topological States}

We construct a probabilistic interpretation of the topological state. For each loop reported by seqPH we compose a mixture of Gaussians that expresses the probability over where this loop is located. The centers of a killer simplex and the centers of the neighboring simplices comprise the means of the components of the mixture. The filtration values associated with each simplex (see Section~\ref{sec:seq_ph}) comprise the weights of the mixture. One natural choice of how to construct the covariances is to treat the 3 vertices $\bv_1,\bv_2,\bv_3 \in \R^3$ of each simplex $s$ as samples from the corresponding Gaussian component $\bx_s \sim \mathcal{N}(\bmu_s, \bSigma_s)$. We can then compute the unbiased sample covariance of these 3 points:
$
\bSigma_s = \tfrac{1}{2} \textstyle\sum_{l=1}^3 (\bv_l - \bmu_s)(\bv_l - \bmu_s)^T
$.
To ensure that $\bSigma$ in non-singular we can place a prior on the covariance and compute posterior treating $\bv_1, \bv_2, \bv_3$ as data. Alternatively, we can use a simpler heuristic of regularizing the covariance with a noise term: $\bSigma_s^{reg} = \bSigma_s + \epsilon \pmb{I}$.
To summarize: our probabilistic interpretation of each loop $l$ that is described by simplices $s_0,...,s_n$ is a Gaussian mixture:
\vspace{-8px}
\begin{equation}
\label{eq:loop_mix}
p_l(\bx) = \textstyle\sum_{i=0}^n w_{s_i} \mathcal{N}(\bx | \bmu_{s_i},\bSigma_{s_i}^{reg})
\end{equation}
where $s_0$ is the killer simplex for the loop $l$ and $s_1,...,s_n$ are the $n$ neighbors of this killer simplex; $w_{s_i}$ is the filtration values of the simplices (the death time in the case of the killer simplex), and $\bmu_{s_i}, \bSigma_{s_i}^{reg}$ are computed based on the  vertices of each simplex $s_i$ as explained above.

\vspace{-5px}
\section{Experiments}

\begin{figure}[t]
\centering
\includegraphics[width=1.0\textwidth]{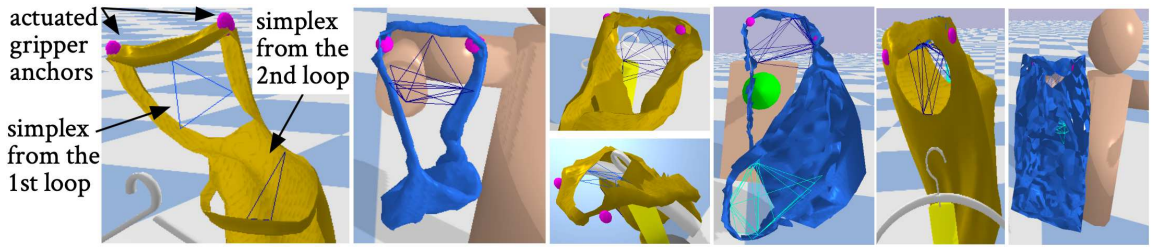}
\vspace{-22px}
\caption{Examples of our scenarios with clothing and bags. Thin lines show killer simplices \& the neighbors.}
\label{fig:sim_all}
\vspace{-10px}
\end{figure}

We created PyBullet~\citep{coumans2019} simulations with objects that have non-trivial topology: clothing items and deformable bags.
At the beginning of each episode two gripper anchors were attached to the deformable object in the scene. They were actuated (with a simple PD controller) to approach the target area with a hanger, a hook or a mannequin figure (Figure~\ref{fig:sim_all}).

For training predictive models we collected 23,000 trajectories, pairing randomly the deformable \& rigid objects in the scene, and randomizing trajectories of the gripper anchors. We also varied elastic and bending stiffness to emulate cloth/deformable materials with various properties.

\begin{figure}[t]
\centering
\includegraphics[width=0.225\textwidth]{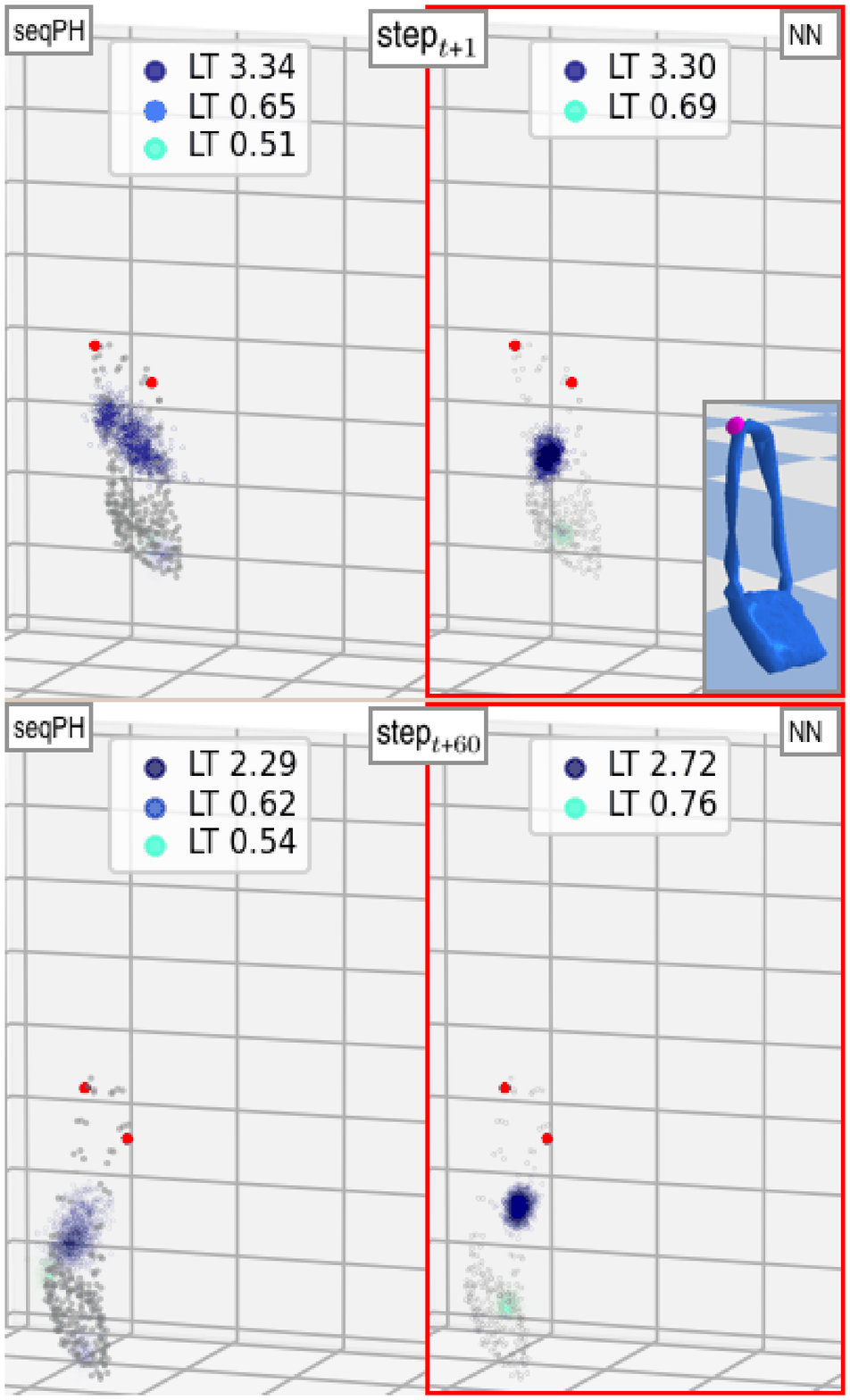}
\includegraphics[width=0.225\textwidth]{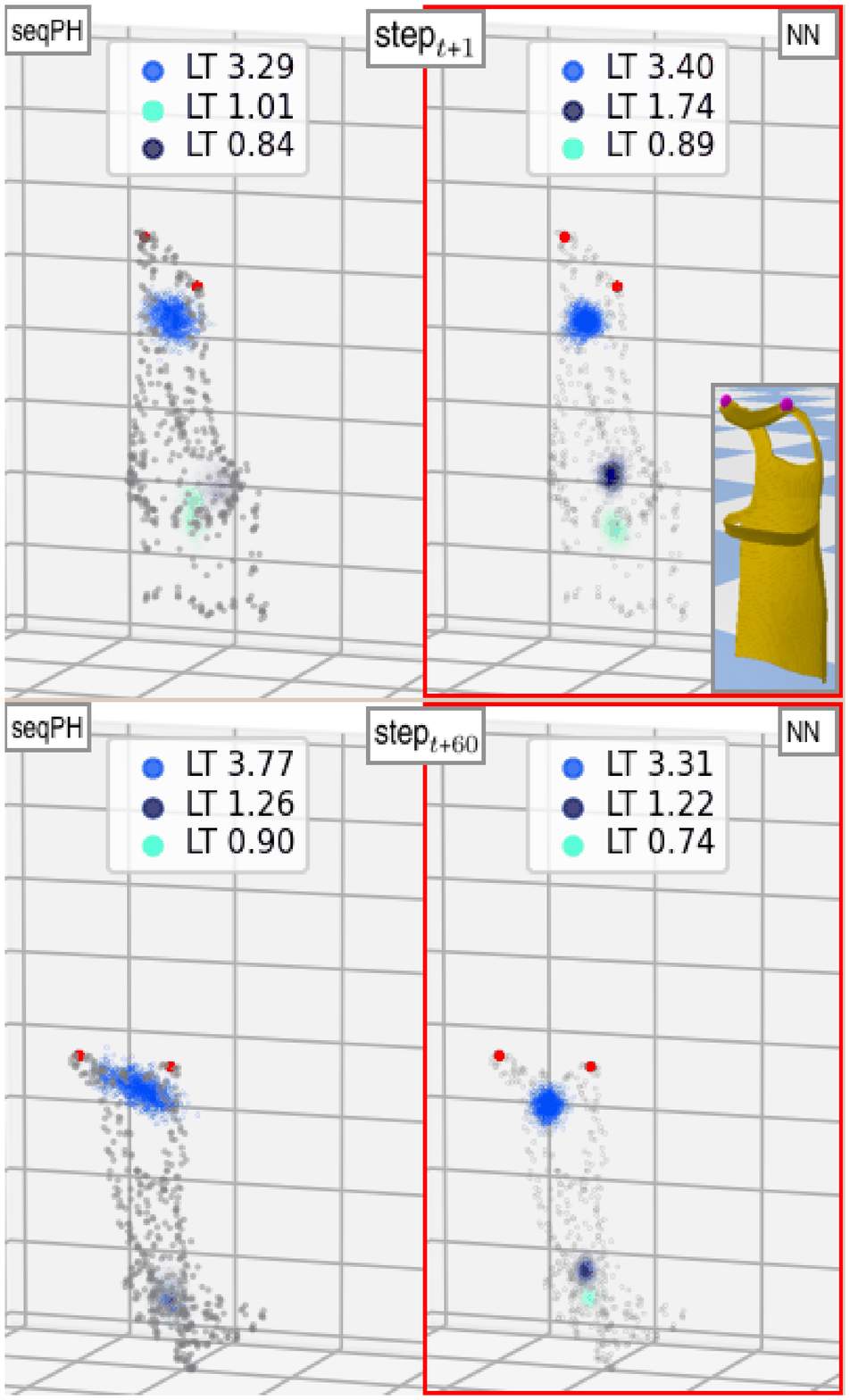}
\includegraphics[width=0.225\textwidth]{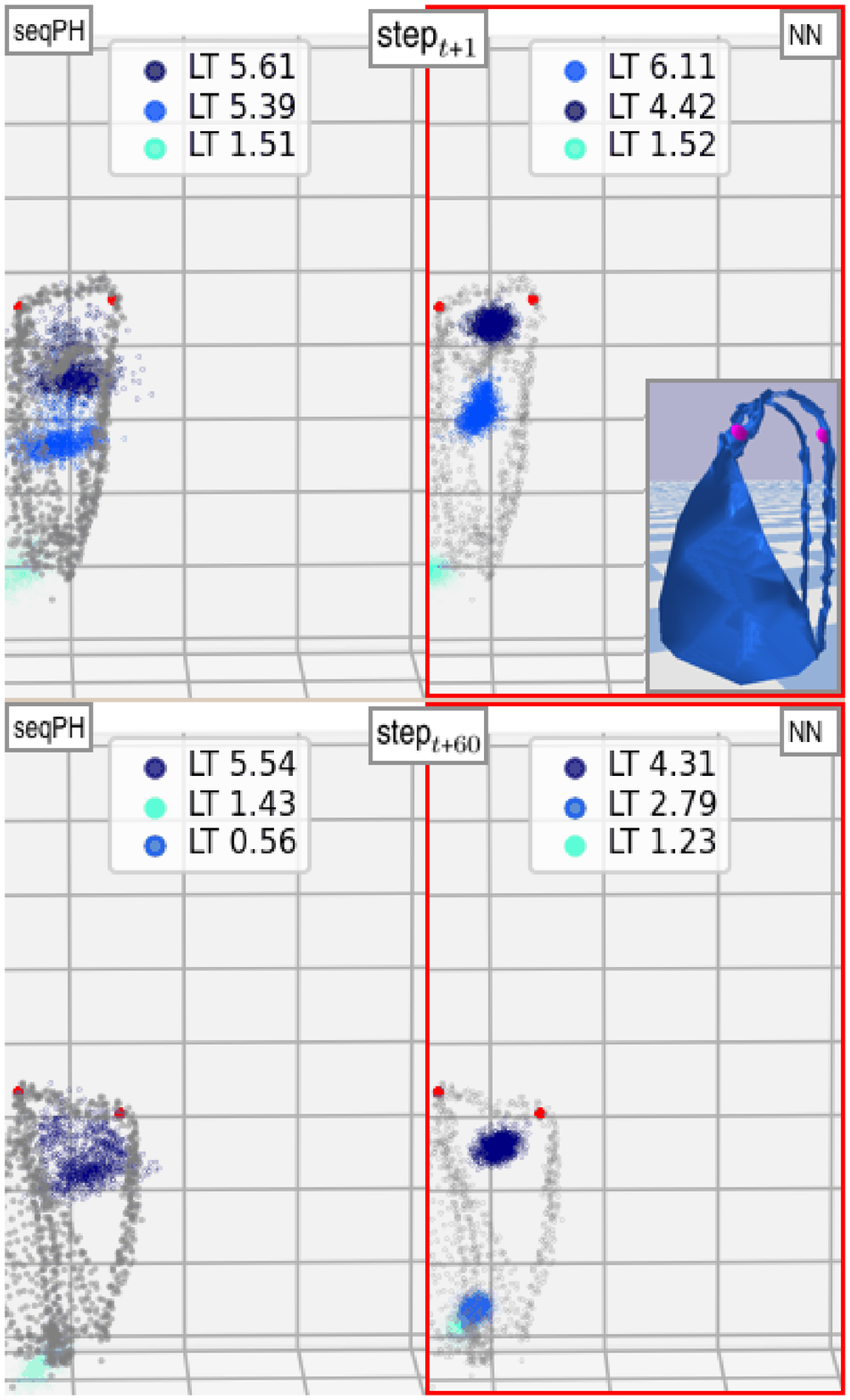}
\vrule
\vspace{1px}
\includegraphics[width=0.29\textwidth]{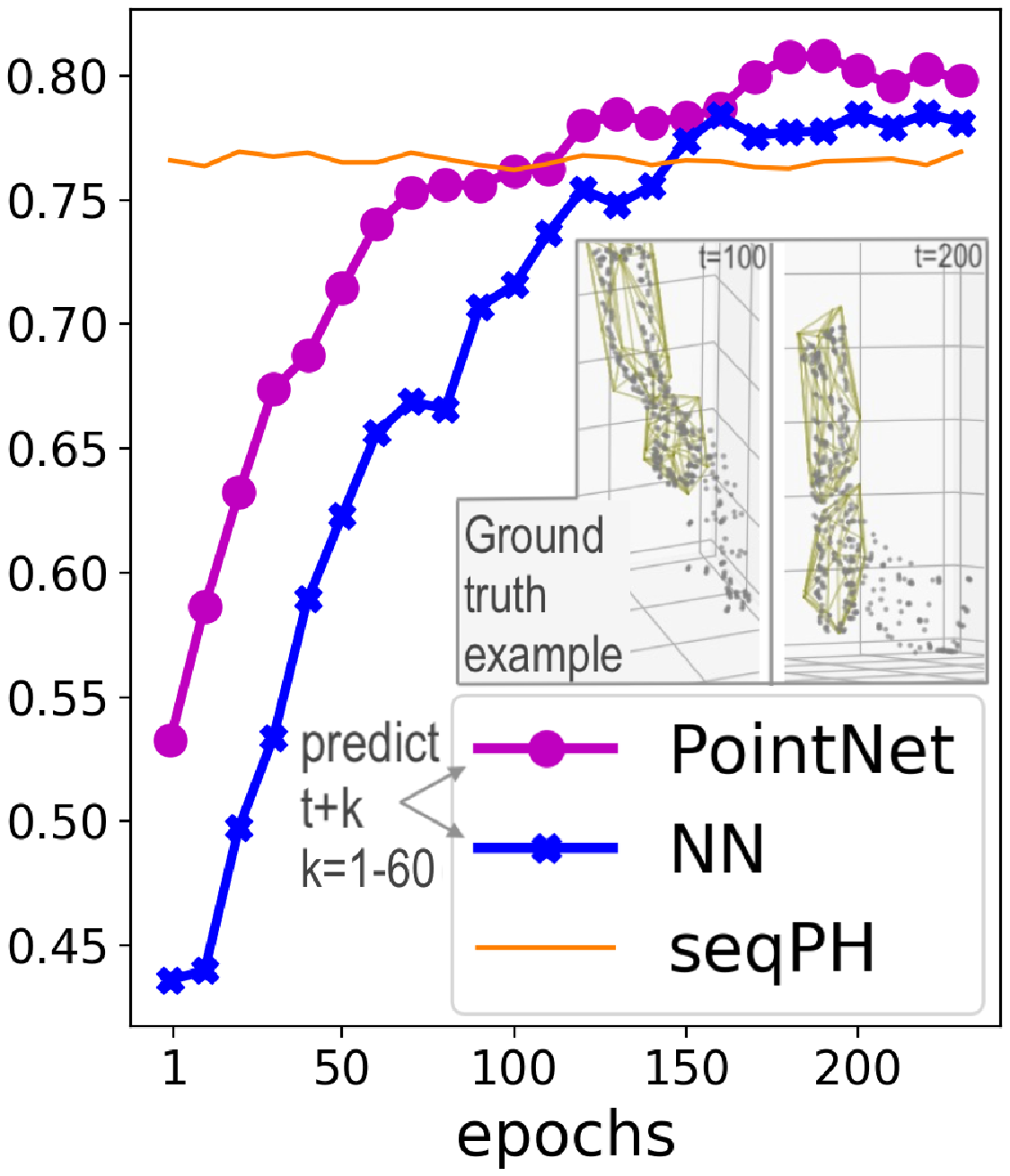}
\vspace{-8px}
\caption{\small{The left sides in each group show results from seqPH. The right sides (with red margins) show mixtures predicted by NN; NN gets point clouds from steps $t\text{-}h\!:\!t$, so the current point cloud is not given as input to NN, it is only visualized for easier interpretation. Left column: results for the small bag object; the dominant mixture/loop (dark blue) has a long lifetime (LT), indicating the loop is large. Middle column: apron (dominant mixture in light blue). Right column: backpack; mixtures for handles in dark \& light blue; the other mixture (in cyan) is phantom, but can be advantageous if it consistently tracks a certain object part. Right plot shows evaluation on a set of objects for which we marked approximate ground truth loop locations.}}
\label{fig:pred_frames}
\vspace{-15px}
\end{figure}

Figure~\ref{fig:pred_frames} illustrates topological states extracted by seqPH (from Section~\ref{sec:seq_ph}) versus those obtained using a predictive NN model (from  Section~\ref{sec:pred_models}).
NN gets as input point clouds from the previous $h\!=\!16$ states (not visualized) and the sequence of proposed actions for the steps $t\!+\!1,...,t\!+\!60$. NN returns predicted topological states for each of the $60$ steps into the future. We visualize predictions for step $t\!+\!1$ and $t\!+\!60$.
Each loop is expressed by a Gaussian mixture, visualized by plotting 1000 samples.
Compared to topological states reported by seqPH, the predictions from NN tend to produce tighter mixtures. This is likely because NN serves as a regularizer, since it is trained on a large number varying trajectories and has to guess the future location of the loop only based on the point clouds from the previous sates and the future/target motion of the gripper anchors. In contrast, topological states extracted from seqPH are based on the point cloud at a given timestep (and loops tracked from the previous steps). Hence, seqPH could be more precise, but could be vulnerable to noise or peculiarities of the current trajectory.

In addition to qualitative evaluation above, we also conducted quantitative evaluation. The latter was highly non-trivial, since the exact ground truth for the topological state was unknown. Hence, we focused on one aspect for which it was tractable to obtain approximate ground truth as follows. We marked a subset of mesh vertices of the main loops in several objects (aprons with one and two loops). Then, we collected a test set of trajectories where these vertices were tracked. To indicate the main area of the loops we computed convex hulls of the tracked points for each loop.
The right plot in Figure~\ref{fig:pred_frames} shows results for the fraction of the test samples (out of $800$) where the mean of the mixture with the longest lifetime was inside the convex hull of the true loop area (for objects with two loops: top two mixtures in two true convex hulls). Predictive model with fully connected architecture (labeled NN) matches results from seqPH. Note that seqPH does not do prediction, we report the loops it extracts from the sequence of the point clouds given to it. The plot also shows that our approach can benefit from the more advanced network architectures, such as PointNet, which outperforms NN and even seqPH results. This demonstrates the ability of NN-based learning to benefit from patterns in the whole dataset and correct the occasional mistakes that seqPH makes. 

\vspace{15px}

\vspace{-5px}
\noindent
\textbf{Conclusion}: We proposed a topological state representation for deformable objects, provided a theoretical formulation for tracking its evolution over time, and designed a method for training a multistep predictive model to enable real-time applications.
The model was tested on scenarios with highly deformable objects and offered fast multistep predictions that improved over both speed and quality of results obtained by employing only computational topology.

\bibliography{references}

\end{document}